\RequirePackage{amsmath}
\documentclass[runningheads]{llncs}
\usepackage{graphicx}

\usepackage[utf8]{inputenc}

\usepackage{amssymb}
\usepackage{wasysym}
\usepackage{scalerel}

\usepackage{mathdots}
\usepackage{subfigure}
\usepackage[abs]{overpic}
\usepackage{nicefrac}
\usepackage[ruled,vlined]{algorithm2e}
\SetKwRepeat{DoWhile}{do}{while}

\usepackage{mathtools}

\DeclareMathOperator*{\argmin}{arg\,min}

\DeclareMathOperator{\dist}{dist}

\DeclareMathOperator{\Log}{Log}
\DeclareMathOperator{\std}{std}

\makeatletter
\newcommand{\ddt}[1][\@nil]{%
  \def\tmp{#1}%
   \ifx\tmp\@nnil
       \frac{\mathrm d}{\mathrm dt}
    \else
        \frac{\mathrm d^#1}{\mathrm dt^#1}
    \fi}
\makeatother

\newcommand{\SO}[1][3]{\text{SO}(#1)}

\newcommand{\Sym}[1][3]{\text{Sym}^+(#1)}

\usepackage[dvipsnames]{xcolor}
\usepackage{color}
\newcommand{\revised}[1]{{#1}}

\newlength{\heightOfB}
\DeclareRobustCommand{\ShowColormap}{\raisebox{-0.14em}{\includegraphics[height=\heightOfB]{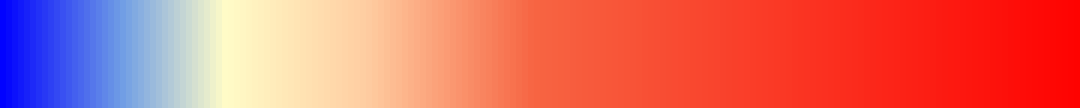}}}

\begin{document}

\title{Geodesic B-Score for Improved Assessment\\ of Knee Osteoarthritis}
%
\author{\revised{Felix Ambellan \and
Stefan Zachow \and
Christoph von Tycowicz}}

%
\authorrunning{\revised{F. Ambellan et al.}}
\titlerunning{\revised{PREPRINT - Geodesic B-Score for Assessment of Knee Osteoarthritis}}

%
\institute{\revised{Visual and Data-centric Computing, Zuse Institute Berlin, Berlin, Germany\\
\email{\{ambellan, zachow, vontycowicz\}@zib.de} }}
\maketitle        

\begin{abstract}

Three-dimensional medical imaging enables detailed understanding of osteoarthritis structural status.
However, there remains a vast need for automatic, thus, reader-independent measures that provide reliable assessment of subject-specific clinical outcomes.
To this end, we derive a consistent generalization of the recently proposed B-score to Riemannian shape spaces.
We further present an algorithmic treatment yielding simple, yet efficient computations allowing for analysis of large shape populations with several thousand samples. 
Our intrinsic formulation exhibits improved discrimination ability over its Euclidean counterpart, which we demonstrate for predictive validity on assessing risks of total knee replacement.
This result highlights the potential of the geodesic B-score to enable improved personalized assessment and stratification for interventions.

\end{abstract}

\section{Introduction}

Osteoarthritis (OA) is a highly prevalent, degenerative joint disease with a considerable societal and economic impact, in addition to the physical and psychological sequelae it causes in affected individuals.
The pathophysiology of OA involves several tissues and is primarily associated with a deterioration of articular cartilage as well as related changes in the underlying bone and at the joint margins.
While OA can affect any joint, knee OA accounts for more than $80\%$ of the global disease burden~\cite{vos2012years}.
There exist various ways of characterizing OA in the literature ranging from subjective assessment to clinical and radiographic ones, albeit with a limited degree of concordance between them.
In practice, plain radiography remains a mainstay for the diagnosis of OA with the Kellgren and Lawrence (KL) grading system~\cite{kellgren1957KLscore} posing the de-facto standard classification scheme.
However, due to its sensitivity on acquisition method and rater reliability, which is reflected in the high number of disagreements between the readers (cf.~\cite{bowes2020machine}), there is a dire need for accurate and reliable assessment of OA status.

Whereas plain radiography only provides 2-dimensional projections, advances in imaging technologies\revised{, especially in magnetic resonance imaging (MRI),} have enabled the understanding of 3-dimensional (3D) OA structural pathology.
In particular, bone shape derived from \revised{MRI} has been found to be associated with radiographic structural progression~\cite{hanik2020splineregression}, to predict radiographic onset of OA~\cite{neogi2013predictOA}, and to discriminate knees w.r.t. osteophyte formation~\cite{tycowicz2020GCN} and OA status~\cite{vonTycowicz2018DCM}.
These findings suggest that bone morphology validly relates to a broader construct of OA pathology.
Furthermore, shape-based assessment holds the promise of reduced sensitivity on image appearance and data acquisition set-ups; e.g.\ systematic changes due to regular technology upgrades with full hardware replacements every 5 to 10 years.
In this light, Bowes et al.~\cite{bowes2020machine} recently introduced a novel, geometrically derived measure to quantify knee OA from bone morphology termed \textit{B-score}.
Contrary to the semi-quantitative KL grade, the B-score is determined fully automatically from femur bone shape and, thus, does not suffer from the subjectivity of the practitioner.
Being a continuous score it enables fine-grained stratification of OA-related structural changes and increases discrimination of risk for clinically important outcomes such as total knee replacement (TKR) surgery.

Despite these recent advances, the formulation of B-score builds upon the popular \textit{active shape model} (ASM)~\cite{Cootes1995ASM} that treats shapes as elements of Euclidean spaces.
However, such linearity assumptions are often inadequate for capturing the high, natural variability in biological shapes (see~\cite{AmbellanLameckervonTycowiczetal.2019} and the references therein).
In particular, sizable empirical improvements can be observed when taking the inherent, geometric structure of shape spaces into account~\cite{davis2010regression,zhang2015shellPCA,ambellan2019FCM,tycowicz2020GCN}.

Since the pioneering work of Kendall~\cite{kendall1989survey}, which introduced a rigorous mathematical foundation for shape spaces and statistics thereon, numerous approaches employing geometric as well as physical concepts such as Hausdorff distance~\cite{charpiat2006distance}, elasticity~\cite{RuWi11,vonTycowicz2015averaging,zhang2015shellPCA}, and viscous flows~\cite{fuchs2009viscousMetric,brandt2016fairing,heeren2018ShellPGA} were devised.
An overview of the various concepts can be found in the chapter by Rumpf and Wirth~\cite{rumpf2015variational}.
Another string of contributions--mainly within the field of computational anatomy--studies shapes in terms of deformations of the ambient space and we refer to~\cite{miller2015hamiltonian} for a comprehensive survey.
In general, these methods suffer from high computational costs and, hence, lack fast response rates limiting their practical applicability especially in large-scale morphological studies. 
To address these challenges, one line of work models shapes by a collection of \revised{elementary building blocks called}  primitives~\revised{(e.g. triangles, M-reps etc.)}~\cite{fletcher2003statistics,freifeld2012LieBodies,vonTycowicz2018DCM,Ambellan2019GLplus} with natural, geometric structure that effectively encodes local changes in shape.
Performing intrinsic calculus on the uncoupled primitives allows for fast computations while, at the same time, accounting for the nonlinearity in shape variation.
Within this category, Ambellan et al.\cite{ambellan2019FCM} recently proposed a surface-theoretic approach that is invariant under Euclidean motion and, thus, 
\revised{is not susceptible to any bias due to misalignment}.

\textbf{Contributions}
In this work, we derive a generalization of the recently proposed B-score to manifold shape spaces that adheres to the rich geometric structure thereof and at the same time is consistent \revised{with} its Euclidean counterpart.
To this end, we build upon a solid mathematical foundation employing concepts from differential geometry and geometric statistics.
We further present an original Newton-type fixed point iteration for projection onto geodesics that is both simple to implement and computationally efficient. 
To the best of our knowledge, previous algorithms restrict to first-order descent schemes~\cite{sommer2014exactPGA} or are tailored to special manifolds~\cite{huckemann2009principal,chakraborty2016CCM-PGA}.
On the application side, we show that the derived geodesic B-score features improved predictive performance on assessing the risk of TKR surgery within 8 years using a single time point.
This result highlights the potential of the Riemannian generalization to enable improved personalized assessment and stratification for interventions.

\section{Background}

\subsection{Shape space}

Before we summarize the employed shape representation, we would like to emphasize that the derived concepts and algorithms provided in this work are not tailored towards a particular choice and are indeed applicable to general Riemannian shape spaces.

For experimental evaluation, we opt for the recently introduced fundamental coordinates model (FCM)~\cite{ambellan2019FCM}.
This model is formulated within the commonly employed deformation-based morphometric framework in which shapes are expressed as deformations of a common reference surface.
More precisely, a digital surface $\mathcal{S}$ is encoded via the orientation preserving deformation $\phi$ of a triangular surface mesh $\bar{\mathcal{S}}$.
For simplicial $\phi$, the deformation gradient $\nabla\phi$ (also known as Jacobian matrix) is a $3\times 3$ matrix of partial derivatives and constant on each triangle of $\bar{\mathcal{S}}$.
In analogy to surface theory, discrete first and second fundamental forms can be derived from $\nabla\phi$ that furnish a complete description of the intrinsic and extrinsic geometry of $\mathcal{S}$.
While the former takes the form of a piece-wise constant (one per triangle) field of $2\times 2$ symmetric positive-definite matrices ($\Sym[2]$), the latter is given by 3D rotations ($\SO$) associated with the edges.
In particular, let $m,n$ be the number of triangles and inner edges, then the resulting shape space is given as the product $\Sigma := \SO^n \times \Sym[2]^m$.
Remarkably, $\Sigma$ can be equipped with a bi-invariant Lie group structure (by virtue of the log-Euclidean framework for $\Sym[2]$~\revised{\cite{pennec2019geomstatsbook,arsigny2006log}}) that lends itself for efficient computations of Riemannian operations.
Furthermore, the FCM provides a Euclidean motion invariant--hence alignment-free--shape representation that assures valid shape instances even in presence of strong nonlinear variability.

\subsection{Geometric statistics}

The nonlinear nature of shape spaces implies that there are no such familiar properties as vector space structure or global system of coordinates (that is, linear combinations of shapes do not generally lie in the space again and shape variations w.r.t. to different base shapes are not directly comparable).
Consequently, core operations pervasive in machine learning and statistics often have to be generalized based on the geometry and specifics of the data at hand.
Approaches that generalize statistical tools to non-Euclidean domains in order to leverage the intrinsic structure belong to the field of geometric statistics and we refer to~\cite{pennec2019geomstatsbook} for an overview.

The simplest--yet also perhaps most fundamentally important--statistic is the sample mean, which estimates the center of a data set.
Because a Riemannian manifold $\mathcal{M}$ has a distance $\dist_\mathcal{M}$ \revised{(length of the shortest path connecting two points)}, we can characterize the mean as the point closest to the data points $x_1,\ldots,x_N \in \mathcal{M}$.
This leads to the notion of (sample) Fr\'echet mean that is the minimizer of the sum-of-squared geodesic distances to the data: 
$$
\mu = \argmin_{x\in\mathcal{M}} \sum_{i=1}^N \dist^2_\mathcal{M}(x, x_i).
$$
While closed-form solutions exist in flat spaces, solving this least-squares problem in general requires iterative optimization routines.
For geodesic manifolds, solutions always exist and are unique for well-localized data~\cite[Thm. 2.1-2.2]{pennec2019geomstatsbook}.

Another fundamental problem is the (statistical) normalization of shape trajectories, i.e. smooth curves in shape space encoding e.g. soft-body motion of anatomical structures.
Normalization of such trajectories into a common reference frame is a challenging task in curved spaces (due to holonomy).
The aim is to preserve as much as possible of the structural variability, while allowing a precise comparison in a common geometric space.
To this end, \textit{parallel transport}~\cite{doCarmo1992} provides a promising approach with a strong mathematical foundation.
Parallel transport allows to propagate a tangent vector (i.e. an infinitesimal shape change) along a path by preserving its properties w.r.t.\ the space geometry, such as a notion of parallelism.

\section{Geodesic B-score}

In this section, we derive a generalization of the recently proposed B-score~\cite{bowes2020machine} to Riemannian shape spaces and present \revised{a} simple, yet effective computational scheme for the determination thereof.
In doing so, our guiding principle is to obtain expressions that take the rich geometric structure of shape space into account (e.g.\ refraining from linearization) and at the same time are consistent \revised{with} its Euclidean counterpart (i.e.\ agree with the original definition for the special case of flat vector spaces).
We term the resulting quantity \textit{geodesic B-score} and will refer to the original definition (respectively, its application in linear spaces) as \textit{Euclidean B-score} whenever this distinction is necessary.

\subsection{Generalization}\label{sec.generalization}
\begin{figure}[t]
\centering
\includegraphics[width=\textwidth]{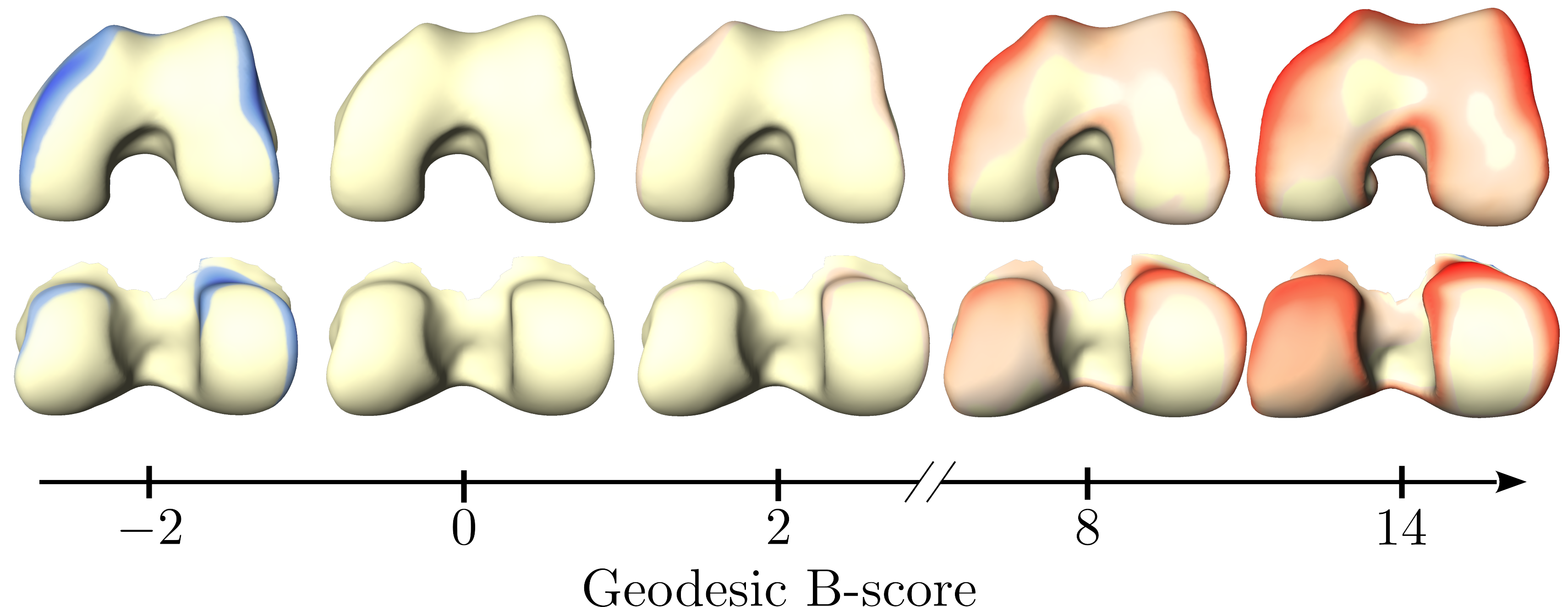}
\caption{Signed vertex deviation from mean shape of mixed-sex non-OA group along the OA-geodesic. \revised{Color-coding}: $-1mm$~\ShowColormap~$5mm$, with neutral window (i.e. yellowish bone color) from $-0.3mm$ to $0.4mm$.
}
\label{fig.OA-geodesic}
\end{figure}

At the core of the construction in~\cite{bowes2020machine} lies \revised{ the projection to an} \textit{OA-vector}
that is defined as the line passing through the mean shapes of populations with and without OA as determined by KL grades $\geq 2$ and $\leq1$, respectively.
While we can readily rely on the Fr\'echet mean, differential geometry provides us with a consistent notion of straight lines known as geodesics~\cite{doCarmo1992}.
In particular, we define the OA-geodesic $\gamma$ as the length minimizing geodesic between the Fr\'echet means of the two populations (which will be unique under the assumptions for the means and the observed overlap of both distributions~\cite{neogi2013predictOA,bowes2020machine}).
A visualization of the OA-geodesic is provided in Fig.~\ref{fig.OA-geodesic} (details on the underlying data are provided in Sec.~\ref{sec.data}).
In order to determine the B-score for a shape $\sigma\in\Sigma$, we first perform an (intrinsic) projection onto the OA-geodesic:
\begin{equation}
    \pi_\gamma(\sigma) := \argmin_{x\in\gamma} \dist^2_\Sigma(x, \sigma).
    \label{eq.proj}
\end{equation}
The signed distance of $\pi_\gamma(\sigma)$ along the OA-geodesic w.r.t.\ the non-OA mean (with positive values in direction of the OA mean) then yields the desired notion of geodesic B-score, i.e.
\begin{align}\label{eq.bscore}
    B_{\gamma,\lambda}(\sigma) &= \lambda \; g_{\gamma(0)}\left(\nicefrac{\dot\gamma(0)}{||\dot\gamma(0)||},\Log_{\gamma(0)} \circ \; \pi_{\gamma}(\sigma)\right),
\end{align}
where $\lambda$ is a positive weighting factor and $g$, $\Log$ denote the Riemannian metric, logarithmic map.
In order to increase interpretability, we take a statistical approach that weights the distances in terms of their distribution within the non-OA population.
More precisely, we employ the Mahalanobis distance such that $\lambda$ is determined as the inverse of the standard deviation $\std(\{B_{\gamma,1}(\sigma) | \sigma\in\mathcal{H}\})$ for the non-OA group $\mathcal{H}$.
In fact, this statistical re-weighting relates the score to the natural morphological inter-subject variability and renders it unitless and scale-invariant.

\subsection{Sex-specific reference}\label{sec.sex}
Females and males have systematically different bone shape~\cite{bowes2020machine} introducing a bias during estimation of the B-score.
\revised{In line with the Euclidean B-score}, we correct for this bias using sex-specific OA-geodesics determined by translating $\gamma$ s.t.\ it passes through the separately computed non-OA mean shapes for each sex.
As a geodesic is uniquely determined by a point and a direction (viz.\ tangent vector at that point), we perform parallel transport of the defining vector along the geodesic connecting the mixed-sex and the sex-specific mean of the respective non-OA group.
\revised{Given sex-specific OA geodesics $\gamma^{{\scaleto{\mars}{6pt}}}, \gamma^{{\scaleto{\venus}{6pt}}}$ we also estimate weighting factors $\lambda^{{\scaleto{\mars}{6pt}}}, \lambda^{{\scaleto{\venus}{6pt}}}$ for each sex and define the B-score as
\begin{align*}
    B(\sigma) &= \begin{dcases}
    B_{\gamma^{{\scaleto{\venus}{5pt}}},\lambda^{{\scaleto{\venus}{5pt}}}}(\sigma) ,& \sigma \text{ female}\\
    B_{\gamma^{{\scaleto{\mars}{5pt}}}, \lambda^{{\scaleto{\mars}{5pt}}}}(\sigma) ,& \sigma  \text{ male}.
\end{dcases}
\end{align*}
}


\subsection{Algorithmic treatment}
Determining solutions to the projection problem in Eq.~(\ref{eq.proj}) does not admit closed-form expressions (except for the special case of constant curvature manifolds~\cite{chakraborty2016CCM-PGA}), thus, requiring iterative optimization.
However, this step is an essential ingredient for the computation of the geodesic B-score.
In order to derive an efficient numerical scheme we assume (without loss of generality) $\gamma : t \mapsto \gamma(t)\in\Sigma$ to be an arc-length parameterized geodesic and express the projection problem as an unconstrained optimization over $t$ with objective function $F(t) = \dist^2_\Sigma(\gamma(t),\sigma)$.
A well-established scheme for this type of problem is Newton's method that employs second-order approximations to gain greatly in convergence speed, achieving quadratic convergence rate when close enough to the optimum.
Analogously, a quadratic approximation for the objective $F$ is given by
\begin{align*}
 F(t+\delta) &\approx F(t) + \ddt F(t) \cdot \delta + \frac{1}{2}\ddt[2] F(t) \cdot \delta^2,~\text{with} \\
 \ddt F(t) &= -2 g_{\gamma(t)}\left(\Log_{\gamma(t)}(\sigma),\dot{\gamma}(t)\right),~\text{and} \\
 \ddt[2] F(t) &= -2 g_{\gamma(t)}\left(\mathrm{d}_{\gamma(t)}\Log_{\gamma(t)}(\sigma)(\dot{\gamma}(t)),\dot{\gamma}(t)\right).
\end{align*}
Additionally, employing the first-order approximation for the differential of the logarithm $\mathrm{d}_{\gamma(t)}\Log_{\gamma(t)}(\sigma) \approx -Id$~\cite[Eq. (5)]{pennec2017hessian} ($Id$ denoting the identity)  
we can obtain an optimal step size $\delta^\ast$ for this quadratic model as
\begin{equation}
    \delta^\ast = g_{\gamma(t)}\left(\Log_{\gamma(t)}(\sigma),\dot{\gamma}(t)\right).
    \label{eg.stepsize}
\end{equation}
Indeed, \revised{verifiable by direct calculation,} this step agrees with the explicit solution for the case of flat spaces.
Eventually, we derive the Newton-type fixed point iteration
\begin{equation}
    \pi_{i+1} = \mathrm{Exp}_{\pi_i}(\delta^\ast_i \dot\gamma_i),
    \label{eq.iteration}
\end{equation}
where $\mathrm{Exp}$ denotes the Riemannian exponential map.
In our setting, the choice $\pi_0 = \gamma(0)$ as initial guess is reasonable, since it is the healthy mean.

\begin{figure}[htb]
\centering
\includegraphics[width=.85\textwidth]{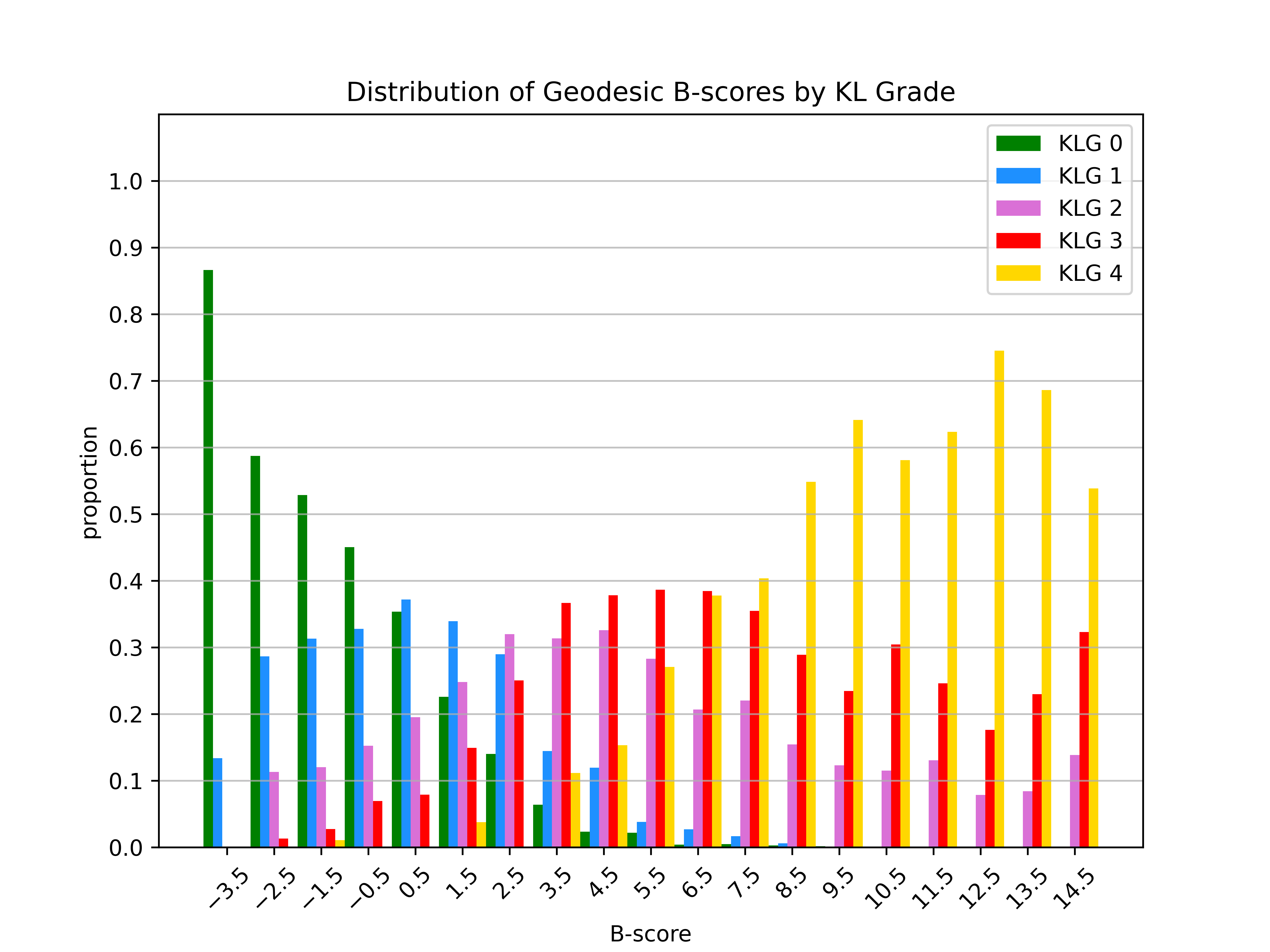}
\caption{Distribution of B-scores by KL grades normalized w.r.t. KL grade imbalance.
}
\label{fig:GeodBscoreByKLNorm}
\end{figure}


\section{Results and discussion}

\subsection{Data description}
\label{sec.data}
Within this practical evaluation we rely on  3D sagittal Double Echo Steady-State \revised{MRI} acquired at baseline as part of the Osteoarthritis Initiative (OAI) database\footnote{nda.nih.gov/oai}\cite{peterfy2006mri}. We segmented the distal femur bone for 9290 (of 9345) scans and established pointwise correspondence employing a fully automatic method combining convolutional neural networks with statistical shape knowledge achieving state-of-the art accuracy~\cite{ambellan-ski10}. The 55 cases not taken into consideration were omitted due to imaging artifacts or failure of the segmentation pipeline. All reconstructed distal femur shapes are taken as left femurs, i.e. all right shapes were mirrored suitably and every shape consists of 8614 vertices and 17024 faces. 
Apart from image data the OAI database also provides clinical scores as KL and information about clinical outcomes such as TKR surgery. An overview on the employed data is given in Table~\ref{table:demographics}. \revised{Note that the list of unique MRI scan IDs defining the study population is available as ancillary file to this manuscript.}
\begin{table}
\centering
\caption{Demographic information for the data under study.}
\label{table:demographics}
\begin{tabular}{l|l}
No. of Shapes & 9290 \\
Laterality (left, right) &  4678, 4612 \\
Sex (male, female) & 3825, 5465  \\
Age [years] & 61.1 $\pm$ 9.2\\
BMI [k/m²] & 28,6 $\pm$ 4.8  \\
KL \revised{0, 1, 2, 3, 4}  & 3382, 1557, 2313, 1198, 282 \\
TKR within 8 years & 508 
\end{tabular}
\end{table}

\noindent{}Since the shape space we employ is not scale invariant (as well as the ASM) this leaves the option to factor it out. However, since femoral osteoarthritis, among others, leads to flattening and widening of the condyle region that at least partially appears as deviation in scale w.r.t. a healthy configuration we forego scale alignment to preserve sensitivity for scale.
Based on the geodesic B-score as derived in Sec.~\ref{sec.generalization}, we restrict our study population to the B-score percentile range from 0.75 to 99.25 (in terms of B-score: -3.12, 14.65) in order to exclude outliers. The resulting distribution of geodesic B-scores per KL grade is shown in Fig.~\ref{fig:GeodBscoreByKLNorm}, \revised{ visualizing the positive correlation of both grading schemes.} Note that the depicted distribution is normalized to account for imbalance within the OAI database of KL grade frequencies\revised{, i.e. re-weighted as if KL groups were of equal cardinality.}


\subsection{Efficiency of projection algorithm}

We empirically evaluate the performance of the derived Newton-type iteration listed in Eq.~\eqref{eq.iteration} using a python-based prototype implementation without parallelization\revised{, publicly available as part of the Morphomatics\footnote{morphomatics.github.io} library}.
To this end, we computed projections of 100 randomly selected femur shapes.
We were able to observe quadratic convergence of the algorithm for all cases with 0.97s and three iterations per case in average.

\subsection{Predictive validity}

\begin{figure}[htb]
\centering \label{fig.tkrrisk}
\includegraphics[width=.8\textwidth]{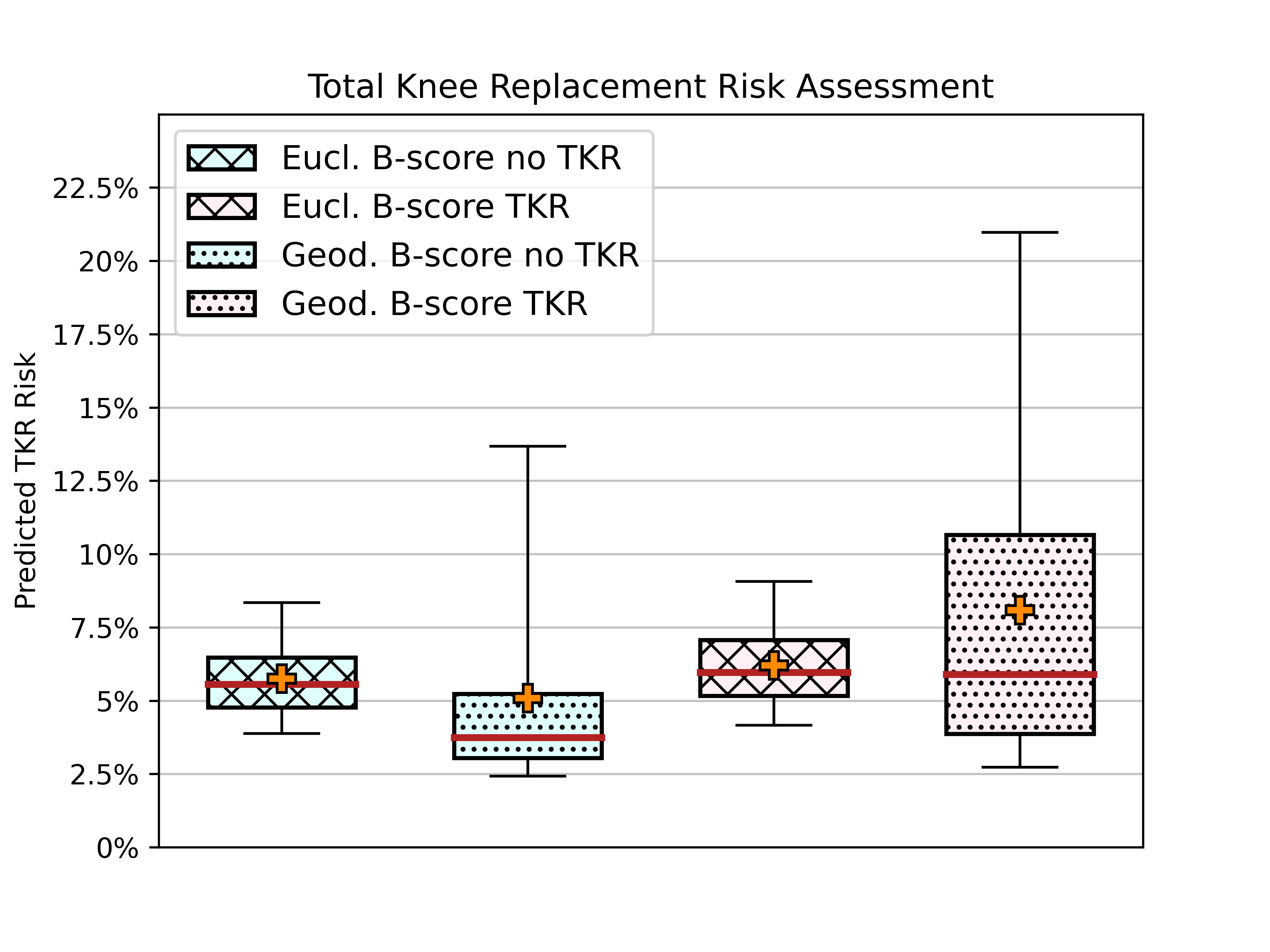}
\caption{Comparison of TKR risk assessment for patient groups with TKR and No TKR clinical outcome respectively. Orange plus: mean, red line: median, box range: 25th to 75th percentile, whisker range: 5th to 95th percentile.
}
\label{fig:TKRrisk}
\end{figure}
We assess the value of the geodesic B-score as a measure of OA status by examining its relationship with risk of TKR surgery---an important clinical outcome.
Here and throughout, we refer to the risk of an outcome as the proportion w.r.t.\ a population.
Additionally, we perform a comparison of the predictive performance between the geodesic and Euclidean B-score.
To this end, we follow the proposed setup from~\cite{bowes2020machine} by modeling the predictor of TKR (within the follow-up period of 8 years) against B-score using logistic regression.

The determination of Euclidean B-scores is based on the space of vertex coordinates.
To reduce confounding effects due to misalignment of the input shapes we employed generalized Procrustes analysis~\cite{Cootes1995ASM} (adding a certain degree of nonlinearity over the approach in~\cite{bowes2020machine}).
No such considerations apply for the FCM-based geodesic B-score as it inherits the invariance to rigid motions.
For both scores, computations were performed on the same input meshes using a modular software design sharing all routines that are not specific to the respective shape space.
To compare the predictive performance of the derived models we grouped the study population into a TKR cohort that did receive TKR and a non-TKR cohort that did not.
In Fig.~\ref{fig:TKRrisk}, we provide box plots for the resulting risk distributions that show clear differences in median risk between non-TKR and TKR.
Furthermore, for the non-TKR cohort the geodesic B-score model validly yields median risks that are half of those for the Euclidean model.
All these differences are statistically significant as determined using Mann--Whitney U tests.
While both approaches yield the same median risk for the TKR cohort, the distribution of the geodesic B-score model is skewed towards higher risks.
These findings substantiate an improved predictive power for the geodesic B-score.

\section{Conclusion and future work}

We introduced a consistent generalization of the recently presented B-score to Riemannian shape spaces.
We showed that the obtained formulation features superior predictive power in an experiment on TKR risk assessment, thus, suggesting improved discrimination of morphological status across the range of OA severity.  
These advances foster the potential of B-score to replace imprecise and insensitive measures for the assessment of OA status based on plain radiography.
Moreover, we further presented an original algorithm for the projection of points in a Riemannian manifold onto a geodesic.
In particular, the obtained iteration exposes fast, quadratic convergence and is simple to implement.

We chose FCM because--due to its deep foundation in differential geometry and link to thin elastic shells--it faithfully captures nonlinear shape variability, while offering fast processing of large-scale shape collections.
On the theoretical side, the price to pay is that there is no guarantee that the projection is diffeomorphic.
However, we would like to remark that the estimated OA-geodesic contains only diffeomorphic deformations within the confidence interval, guaranteeing valid instances even if the input shapes are not.
Furthermore, contrary to shape spaces based on diffeomorphic metric mapping, the FCM is invariant under Euclidean motion and, thus, not susceptible to any bias due to misalignment.

In this work, we carefully generalized the B-score mimicking the geometric construction of the Euclidean counterpart.
However, there are various statistical approaches that allow to estimate submanifolds based on separation or regression considerations, e.g. geodesic discriminant analysis~\cite{louis2018GDA} or higher-order regression~\cite{hanik2020splineregression}, respectively.    
An interesting direction for future work is to investigate to which extend such geometric statistics can serve as a foundation for advanced notions of an intrinsic B-score.
From a medical perspective, it will be most interesting to explore the relationship of the geodesic B-score with further clinically important outcomes such as pain and loss of function.
In particular, we will investigate to which degree the geodesic B-score can improve the related risk assessment. \revised{Since the presented statistical approach can directly be extended to multiple connected components, another line of work will aim on extension to multi-structure B-scores, e.g. for Femur and Tibia.}
Moreover, for the future we envision a longitudinal characterization beyond the static B-score that takes subject-specific shape developments into account.
\section*{Acknowledgments}
\revised{F. Ambellan is funded by the Deutsche Forschungsgemeinschaft (DFG, German Research 
Foundation) under Germany's Excellence Strategy – The Berlin Mathematics 
Research Center MATH+ (EXC-2046/1, project ID: 390685689). This work was supported by the Bundesministerium fuer Bildung und Forschung (BMBF) through BIFOLD - The Berlin Institute for the Foundations of Learning and Data (ref. 01IS18025A and ref 01IS18037A).}
Furthermore, we are grateful for the open-access dataset of the Osteoarthritis Initiative\footnote{Osteoarthritis Initiative is
a public-private partnership comprised of five contracts
(N01-AR-2-2258; N01-AR-2-2259; N01-AR-2-2260; N01-AR-2-2261; N01-AR-2-2262) 
funded by the National Institutes of Health, a branch of the Department of Health and 
Human Services, and conducted by the OAI Study Investigators. Private funding partners 
include Merck Research Laboratories; Novartis Pharmaceuticals Corporation, 
GlaxoSmithKline; and Pfizer, Inc. Private sector funding for the OAI is managed by the 
Foundation for the National Institutes of Health. This manuscript was prepared using an OAI 
public use data set and does not necessarily reflect the opinions or views of the OAI 
investigators, the NIH, or the private funding partners.}.

\bibliographystyle{splncs04}
\bibliography{geodesicBscore}

\end{document}